\documentclass[journal]{IEEEtai}

\usepackage[colorlinks,urlcolor=blue,linkcolor=blue,citecolor=blue]{hyperref}
\usepackage{color,array}
\usepackage{graphicx}
\usepackage{amsmath,amssymb}
\usepackage{booktabs}
\usepackage{float}
\usepackage{listings}
\usepackage{caption}
\usepackage{algorithm}
\usepackage{algpseudocode}

\graphicspath{{../analysis_output_v2/final graphs/}{./}}

\begin{document}

\title{Enhanced LLM Reasoning by Optimizing Reward Functions with Search-Driven Reinforcement Learning}

\author{Arash Ahmadi, Sarah Sharif, Yaser (Mike) Banad
\thanks{A. Ahmadi, S. Sharif, and Y. M. Banad are with the School of Electrical and Computer Engineering, University of Oklahoma, Norman, OK 73019 USA (e-mail: arash.ahmadi@ou.edu; s.sh@ou.edu; bana@ou.edu). All authors are members of the Intelligent Neuromorphic and Quantum Understanding for Innovative Research and Engineering (INQUIRE) Laboratory, University of Oklahoma, Norman, OK, USA. \textit{Corresponding author:} Y. M. Banad.}\thanks{Code, generated reward functions, and experimental artifacts are available at \url{https://github.com/INQUIRELAB/search-reward-rl}.}}

\maketitle

\begin{abstract}
Mathematical reasoning is a key benchmark for large language models. Reinforcement learning is a standard post-training mechanism for improving the reasoning capabilities of large language models, yet performance remains sensitive to the design of the reward function that drives policy optimization. This paper introduces a search-driven framework that treats the reward specification itself as an object of optimization. The setting of interest is one in which the base model is held fixed and the reward specification is the primary remaining design lever. Candidate reward functions are generated by a frontier language model, validated automatically, screened through 500-step Group Relative Policy Optimization (GRPO) training runs on a Llama-3.2-3B-Instruct base model with Low-Rank Adaptation (LoRA), and ranked by F1 on the GSM8K test set. Ranked summaries from prior rounds are then fed back into the next round of generation. Over five rounds, the search produces 50 candidate rewards. The mean F1 rises from 0.596 in Round~1 to 0.632 in Round~5, and the top individual reward reaches F1\,=\,0.787. Seven ensemble configurations of top-ranked rewards are evaluated. The best ensemble achieves F1\,=\,0.795 (95\% bootstrap CI [0.756,\,0.832]) and accuracy 0.660 [0.635,\,0.686], a 0.19 absolute F1 gain over a base-rewards-only GRPO baseline (F1\,=\,0.609). Pairwise McNemar tests with Bonferroni correction show all five-or-more-reward configurations are statistically indistinguishable at $\alpha\,=\,0.05/21$. A three-seed re-training of the best ensemble yields F1 of $0.785\pm 0.013$. A randomly drawn 5-reward control collapses to F1\,=\,0.047, which shows that the ranked-feedback loop, not the additive signal of having more rewards, drives the gain.
\end{abstract}

\begin{IEEEImpStatement}
Reinforcement learning is increasingly used to improve reasoning in AI language models, but designing effective reward functions remains a manual and error-prone process that typically requires expert intuition and trial-and-error. The framework presented here automates reward discovery and reduces human design effort while keeping each reward as readable Python code that practitioners can audit, modify, and combine. The pipeline runs on a single consumer-grade GPU in roughly 40 hours, which puts the method inside the reach of academic and small-industry compute budgets. The discovered rewards measurably improve mathematical-reasoning quality on a fixed base model and produce fewer incorrect answers when the model does respond. The same recipe transfers to other reasoning domains such as code generation and scientific problem solving and offers a practical alternative to expensive reward-model training for aligning language models with target reasoning behaviors. All code and discovered reward functions are released at \url{https://github.com/INQUIRELAB/search-reward-rl}.
\end{IEEEImpStatement}


\section{Introduction}

\IEEEPARstart{M}{athematical} reasoning has become one of the clearest stress tests for large language models (LLMs) because success requires more than fluent text generation. A capable model must preserve intermediate consistency, carry quantities over multiple steps, and produce a final answer that is both correct and precisely stated. Prompting-based methods such as chain-of-thought prompting and self-consistency improved arithmetic and symbolic reasoning substantially, and benchmarks such as GSM8K made these gains measurable at scale~\cite{wei2022chain,wang2022self,cobbe2021training}. However, prompting changes how a model is queried rather than what the model is rewarded to do during training.

Reinforcement learning (RL) has re-emerged as a central post-training mechanism for aligning LLMs with desired behaviors. Modern policy optimization methods such as Proximal Policy Optimization (PPO) made RL stable enough for large-scale language applications~\cite{schulman2017proximal}. In language modeling, the same logic underlies reinforcement learning from human feedback (RLHF), where human or model-derived preferences are converted into training signals that push a policy beyond imitation of static demonstrations~\cite{christiano2017deep,ouyang2022training,bai2022constitutional}. Direct Preference Optimization (DPO) shows that preference optimization can sometimes bypass explicit online RL, but it does not remove the underlying question of what constitutes a useful reward or preference signal~\cite{rafailov2023direct}.

That question becomes especially important in reasoning tasks. A reward that scores only the final answer is easy to compute when ground-truth labels are available, but it is also coarse: it cannot reliably distinguish a correct answer reached through faithful reasoning from one reached by shallow pattern matching. A reward that evaluates intermediate reasoning steps offers richer credit assignment, yet it is much harder to design, annotate, and validate at scale~\cite{lightman2023let,zhang2025lessons}. Reward models themselves vary in quality and robustness~\cite{lambert2403rewardbench}. Reasoning performance is often limited not only by the base model or the policy optimizer, but also by the quality of the reward specification. The broader RL literature has long recognized that reward design is a bottleneck. Human-written rewards are often proxies for the behavior a practitioner actually wants, and poorly chosen proxies can induce reward hacking, shortcut learning, or brittle behavior~\cite{hadfield2017inverse}. This has motivated work on intrinsic rewards, meta-learned objectives, and automated RL systems that treat the objective itself as something to be optimized rather than assumed fixed~\cite{zheng2018learning,houthooft2018evolved,parker2022automated}.

Recent advances in LLMs have made reward design more programmable. Modern language models can synthesize, critique, and revise executable code, and they can participate directly in reward construction. Prior work has used LLMs as proxy reward functions, transformed language instructions into reward parameters, generated dense reward code from task descriptions, and iteratively improved reward programs through performance feedback~\cite{kwon2023reward,yu2023language,xie2023text2reward,ma2023eureka,hazra2024revolve,sun2025large}. The common lesson is that reward design can be partially outsourced to a generative model when it is embedded in an iterative feedback loop. In parallel, reasoning-focused post-training has shown that RL can materially improve mathematical and logical reasoning. Group Relative Policy Optimization (GRPO), introduced for DeepSeekMath, replaces the separate critic used in PPO with group-relative normalization of sampled outputs. This change reduces memory overhead and improves mathematical reasoning~\cite{shao2024deepseekmath}. DeepSeek-R1 further demonstrated that RL can elicit stronger reasoning behaviors such as self-correction at scale~\cite{guo2025deepseek}. Yet in most of this literature the reward is still treated as fixed infrastructure: a handcrafted rule set, a trained reward model, or a single verifier.

This paper studies a complementary question: whether the reward specification used during RL post-training can itself be the object of search. The setting of interest is one in which the base model is held fixed by the practitioner and the reward function becomes the dominant remaining design variable. The framework introduced here treats the reward function as an object of search. Candidate reward functions are generated, validated, screened through GRPO training runs, ranked by downstream reasoning performance, and then fed back into the next round of reward generation. Rather than asking only which policy performs best under a fixed reward, this work asks whether the reward can itself be progressively improved as a sequence of competing hypotheses about what good mathematical reasoning looks like.

The research makes four contributions. First, it formulates reward search for language-model reasoning as an iterative generate--validate--train--rank loop over executable reward functions. Second, it applies this loop under GRPO, a reasoning-relevant and computationally efficient policy optimizer. Third, it examines whether multiple discovered rewards can be combined into stronger ensemble signals than any individual reward alone. Fourth, it provides an empirical analysis, including bootstrap confidence intervals, pairwise McNemar tests with multiple-comparison correction, and a reward-hacking audit, of how the discovered rewards trade off reasoning quality, format compliance, and computational cost on GSM8K.

\section{Related Work}

\subsection{Mathematical Reasoning in LLMs}

Mathematical reasoning became a focal evaluation domain for LLMs because it exposes failures that remain hidden in more open-ended benchmarks. GSM8K provided a standardized set of grade-school word problems and became a leading testbed for multi-step arithmetic reasoning~\cite{cobbe2021training}. Chain-of-thought prompting encouraged models to externalize intermediate reasoning steps, while self-consistency improved answer selection by aggregating over multiple sampled reasoning paths~\cite{wei2022chain,wang2022self}. Verifier-based reranking improved answer accuracy by selecting among candidate solutions, but it treated reasoning quality as something to evaluate after generation rather than something to optimize directly during policy learning~\cite{cobbe2021training}.

\subsection{Preference-Based Alignment and Reward Models}

Preference-based alignment changed how reward is obtained rather than eliminating the reward problem. Christiano et al. showed that pairwise human comparisons can supervise agents without access to the ground-truth reward function~\cite{christiano2017deep}. InstructGPT translated this into a pipeline of supervised fine-tuning, reward-model training, and PPO-based policy optimization~\cite{ouyang2022training}. Constitutional AI reduced reliance on direct human labels by using generated critiques~\cite{bai2022constitutional}. DPO argued that preference optimization can recover the same target as standard RLHF without explicit online RL~\cite{rafailov2023direct}. RewardBench made the fragility of this approach visible by showing that reward models differ in robustness over chat, reasoning, and safety tasks~\cite{lambert2403rewardbench}. Self-Rewarding Language Models push the idea further by letting a language model generate its own preference-style supervision during iterative optimization~\cite{yuan2024self}.

\subsection{Outcome and Process Supervision}

The main reward-design distinction in reasoning tasks is between outcome supervision and process supervision. Outcome supervision evaluates the final answer as a whole, and verifier-based methods for math reasoning have shown that even coarse outcome judgments can materially improve performance~\cite{cobbe2021training}. Process supervision evaluates intermediate reasoning steps. Lightman et al. argue that step-level feedback produces more reliable reward models than outcome supervision alone~\cite{lightman2023let}. Zhang et al. show that building strong process reward models is itself difficult: data synthesis choices and annotation quality can distort conclusions about true reasoning quality~\cite{zhang2025lessons}. A search over reward functions can target final correctness, structural formatting, step quality, or mixtures, thereby occupying a middle ground between pure outcome rewards and fully trained process reward models.

\subsection{Automating Reward Design}

Before LLMs were used to write reward code, RL researchers had begun exploring automated objective design. Zheng et al. learned intrinsic rewards that accelerated policy-gradient training~\cite{zheng2018learning}. Houthooft et al. evolved differentiable loss functions for policy learning~\cite{houthooft2018evolved}. The AutoRL literature generalized this view by treating reward design, algorithm choice, and hyperparameter tuning as automatable components of a broader RL pipeline~\cite{parker2022automated}. EAGER derived auxiliary rewards from natural-language goals by asking and answering questions about the desired task~\cite{carta2022eager}.

Modern LLMs made automated reward design more accessible. Reward Design with Language Models used a prompted LLM as a proxy reward evaluator~\cite{kwon2023reward}. Language to Rewards translated language instructions into reward parameters for robotic skill synthesis~\cite{yu2023language}. Text2Reward generated dense reward programs from task descriptions, while Eureka used a coding LLM to iteratively improve reward code that in many cases matched expert-designed rewards~\cite{xie2023text2reward,ma2023eureka}. REvolve framed reward construction as an evolutionary process with populations, mutation, crossover, and selection guided by feedback~\cite{hazra2024revolve}. CARD reduced human burden further by using dynamic feedback to improve reward code iteratively~\cite{sun2025large}. However, the strongest empirical evidence remains concentrated in robotics and control. This paper transfers the core idea of iterative reward generation and selection into mathematical reasoning for language models.

\subsection{GRPO and Reasoning-Centered RL}

GRPO replaces the separate critic used in PPO with group-relative normalization of sampled outputs and thereby reduces memory overhead while improving mathematical reasoning~\cite{shao2024deepseekmath}. DeepSeek-R1 extended this trend by showing that RL can induce stronger reasoning behavior and self-correction in open models~\cite{guo2025deepseek}. Logic-RL demonstrates that rule-based rewards and strict formatting constraints can stabilize reasoning-oriented RL even in small-data settings~\cite{xie2025logic}. LoRA and related parameter-efficient fine-tuning methods play an enabling role by making repeated adaptation feasible without full-model retraining~\cite{hu2022lora}. Despite this progress, most reasoning-focused RL pipelines still assume that the reward function is already known.

\section{Methods}
\label{sec:methods}

\begin{figure*}[p]
\centering
\includegraphics[width=\textwidth,height=0.92\textheight,keepaspectratio]{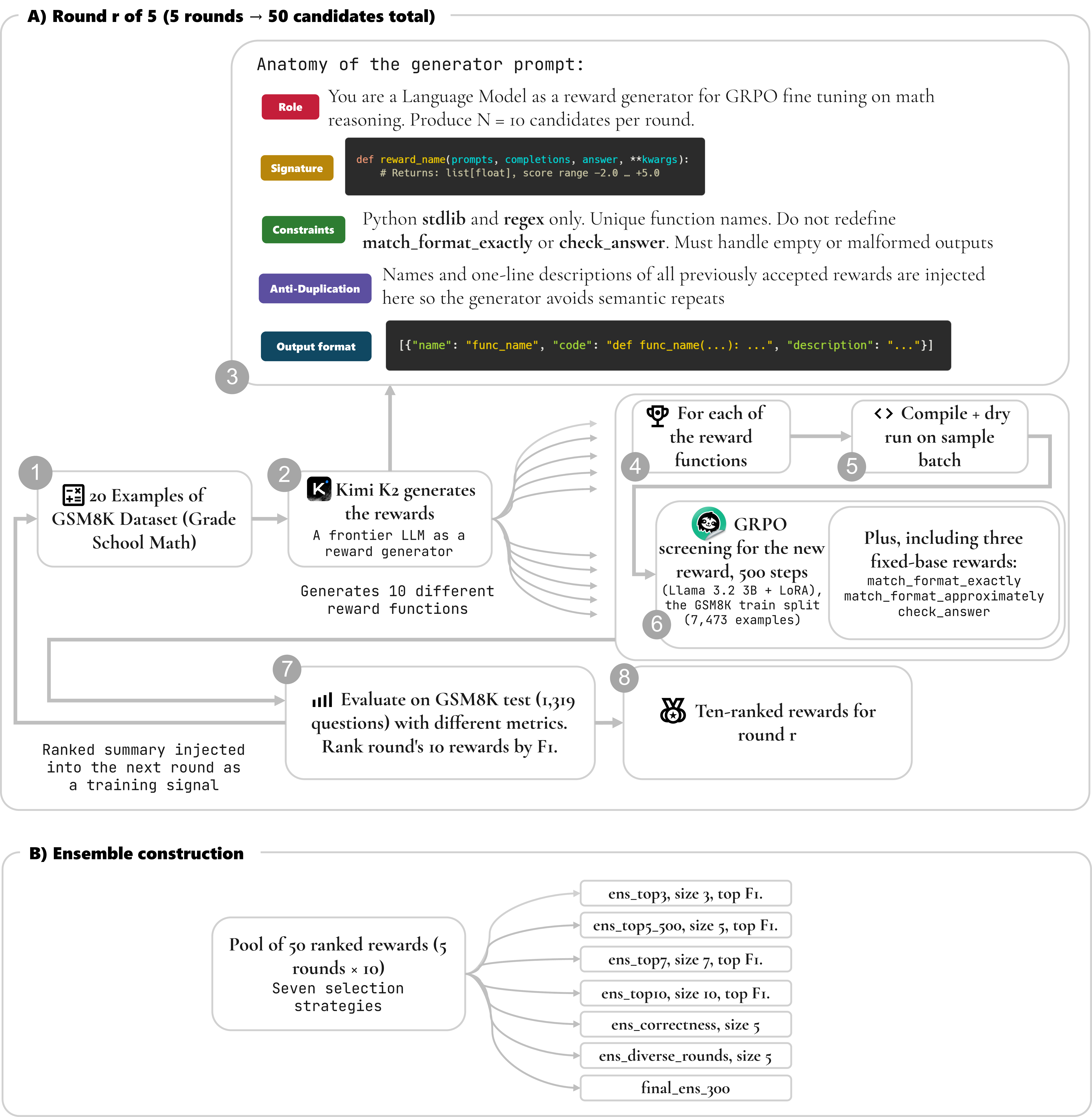}
\caption{Framework overview. Panel~(A) details one round of search-driven reward synthesis (of five). Twenty GSM8K examples and the anatomy of the generator prompt, role, required function signature, constraints (stdlib only, no override of base rewards, graceful handling of malformed outputs), anti-duplication block that injects names and descriptions of all previously accepted rewards, and the expected JSON output format, are passed to Kimi K2, which produces ten candidate reward functions. Each candidate is compiled and dry-run, then screened for 500 GRPO steps on Llama-3.2-3B-Instruct with LoRA on the GSM8K train split alongside the three fixed base rewards, evaluated on the 1{,}319-question test split, and ranked by F1. The ranked summary is injected into the next round as the feedback signal. Panel~(B) shows ensemble construction from the pool of 50 ranked rewards into seven selection strategies.}
\label{fig:architecture}
\end{figure*}

The proposed framework consists of three major components: the GRPO reinforcement learning algorithm, a search-driven feedback strategy that uses a frontier LLM to generate and iteratively refine reward functions, and an ensemble mechanism that combines top-performing reward functions into a unified training signal. Fig.~\ref{fig:architecture} provides an overview of the full pipeline.

\subsection{Group Relative Policy Optimization}

Given an input question $q$, GRPO samples a group of $G$ outputs $\{o_1, o_2, \ldots, o_G\}$ from the old policy $\pi_{\theta_{\text{old}}}$. For each output $o_i$, a reward $r_i$ is computed. Under outcome supervision, the rewards are normalized by subtracting the group mean and dividing by the group standard deviation, and the resulting normalized value is assigned as the advantage of every token in the output~\cite{shao2024deepseekmath}:
\begin{equation}
\hat{A}_{i,t} = \widetilde{r}_i = \frac{r_i - \operatorname{mean}(\mathbf{r})}{\operatorname{std}(\mathbf{r})}
\label{eq:grpo_advantage}
\end{equation}
where $\mathbf{r} = \{r_1, r_2, \ldots, r_G\}$ is the vector of group rewards. The group-relative baseline replaces the learned value function used in PPO.

The GRPO training objective maximizes the clipped surrogate loss:
\begin{equation}
\begin{split}
\mathcal{J}_{\text{GRPO}}(\theta) = {} & \mathbb{E}_{q,\{o_i\}} \Bigg[\, \frac{1}{G} \sum_{i=1}^{G} \frac{1}{|o_i|} \sum_{t=1}^{|o_i|} \\
& \Big\{ \min\!\big( \rho_{i,t}\hat{A}_i,\; \operatorname{clip}(\rho_{i,t}, 1{-}\varepsilon, 1{+}\varepsilon)\hat{A}_i \big) \\
& - \beta\, D_{\text{KL}}\!\left[\pi_\theta \,\|\, \pi_{\text{ref}}\right] \Big\} \Bigg]
\end{split}
\label{eq:grpo_objective}
\end{equation}
where $\rho_{i,t} = \pi_\theta(o_{i,t}\mid q, o_{i,<t}) / \pi_{\theta_{\text{old}}}(o_{i,t}\mid q, o_{i,<t})$ is the probability ratio between the current and old policy, $\varepsilon$ is the clipping parameter, and $\beta D_{\text{KL}}$ regularizes against the reference policy to prevent reward hacking. Following DeepSeekMath, the KL term sits inside the per-token sum and uses the unbiased $k_3$ estimator $\pi_{\text{ref}}/\pi_\theta - \log(\pi_{\text{ref}}/\pi_\theta) - 1$ rather than an outer penalty~\cite{shao2024deepseekmath}.

The elimination of the critic network reduces memory footprint by approximately half, removes the need to train a separate value function to convergence, and avoids staleness issues that arise when a separately trained critic lags behind the improving policy. These properties make GRPO well-suited for parameter-efficient fine-tuning on limited hardware.

\subsection{Base Model and LoRA Configuration}

The base model is Meta's Llama-3.2-3B-Instruct~\cite{grattafiori2024llama}, loaded in 4-bit quantized precision using bitsandbytes. Low-Rank Adaptation (LoRA) decomposes weight updates into low-rank matrices:
\begin{equation}
W' = W + BA
\label{eq:lora}
\end{equation}
where $W$ is the frozen pre-trained weight matrix, and $B \in \mathbb{R}^{d \times r}$, $A \in \mathbb{R}^{r \times k}$ are the trainable low-rank matrices with rank $r = 32$. LoRA adapters are attached to all attention and feed-forward projection matrices (q, k, v, o, gate, up, down). This results in approximately 83 million trainable parameters out of 3 billion total (2.8\%). Each reward function trial saves its own LoRA checkpoint, and after evaluation the adapter is unloaded so that no learned behavior leaks between trials.

\subsection{Dataset}

The training and evaluation dataset is GSM8K, which contains 7,473 training examples and 1,319 test examples~\cite{cobbe2021training}. All GRPO training uses only the training split. All evaluation metrics are computed on the full 1,319-question test set.

The model is expected to produce output in a structured format:
\begin{verbatim}
<thinking>[Reasoning]</thinking>
<solution>[Final numeric answer]</solution>
\end{verbatim}
This format separates the reasoning process from the final answer so that reward functions can independently assess reasoning quality and answer correctness.

\subsection{Base Reward Functions}
\label{sec:base-rewards}

Three fixed base reward functions are included in every training run, both during reward screening and in all ensemble configurations. They are declared in the codebase as
\begin{verbatim}
BASE_REWARD_NAMES = [
 "match_format_exactly",
 "match_format_approximately",
 "check_answer",
]
\end{verbatim}
and are protected from being overridden or replaced by any generated reward; the four-stage sandbox rejects any candidate that attempts to redefine them. They form the always-on substrate against which discovered rewards are tested. The first, \texttt{match\_format\_exactly}, checks whether the output contains properly paired \texttt{<thinking>} and \texttt{<solution>} tags and awards $+3.0$ when matched and $0$ otherwise. The second, \texttt{match\_format\_approximately}, provides a graduated formatting signal: $+0.5$ for each of the four tags present exactly once, or $-1.0$ for each missing or duplicated tag. The third, \texttt{check\_answer}, evaluates the extracted numeric answer against ground truth with tiered scoring: $+3.0$ for exact match, $+1.5$ after whitespace normalization, $+0.9$ for answers in a $\pm 10\%$ ratio, $+0.5$ for a $\pm 20\%$ ratio, and $-1.5$ for wrong answers; missing or unparseable answers receive $-2.5$. The combined reward used during any GRPO trial of a candidate $R$ is therefore $R_{\text{train}}(o)= R^{\text{exact}}(o)+R^{\text{approx}}(o)+R^{\text{check}}(o)+R(o)$, which means every discovered reward is always evaluated as an additive contribution on top of the three base rewards rather than as a replacement.

\subsection{Search-Driven Reward Function Generation}
\label{sec:methods-iterative}

The core contribution is a search-driven feedback loop that discovers reward functions beyond what the three base rewards can achieve. The process proceeds over five rounds, each generating 10 new candidate reward functions (50 total).

\subsubsection{Reward Generation via LLM}

Reward functions are generated by prompting Kimi K2 (\texttt{moonshotai/kimi-k2-instruct-0905}) through the Groq API. The generation prompt provides: (1) a random sample of 20 training examples, each consisting of the question and the ground-truth final answer; (2) an example reward function demonstrating the expected signature; (3) a specification requiring each function to accept \texttt{prompts}, \texttt{completions}, and \texttt{answer} as arguments and return a list of floats; and (4) a constraint that function names must not duplicate previously generated ones nor any of \texttt{match\_format\_exactly}, \texttt{match\_format\_approximately}, or \texttt{check\_answer} (the protected base rewards from Section~\ref{sec:base-rewards}). Generation uses temperature 0.6 with a maximum of 8,192 output tokens. The model returns a JSON array of name, code, and description fields. Each candidate that passes validation is then trained with GRPO on top of the three fixed base rewards for 500 steps using the hyperparameters described in Section~\ref{sec:training}, and evaluated on the GSM8K test set to obtain the F1, accuracy, precision, and recall used to rank rewards within and across rounds.

\subsubsection{Feedback Loop}

After each round, all previously evaluated reward functions are ranked by F1 score on the test set. A ranked summary of these results is passed back to Kimi K2 in the prompt for the next round, explicitly instructing the model to avoid duplicating names or semantics and to focus on uncovered aspects of mathematical reasoning. This ranked-feedback mechanism supplies implicit selection pressure: high-scoring rewards survive as descriptive information that conditions the next round of generation, while lower-scoring ones implicitly signal directions to avoid. The signal re-entering the next round is a compact textual summary of the form ``\texttt{name=$\langle$reward$\rangle$, round=$\langle$r$\rangle$, F1=$\langle$value$\rangle$, acc=$\langle$value$\rangle$}'', and the search space itself has no learnable parameters.

\subsubsection{Validation and Screening}

Each generated reward function passes through a four-stage validation pipeline: (1) an AST pass rejects imports outside a small standard-library whitelist; (2) the code is executed in a locked-down environment with stripped \texttt{\_\_builtins\_\_}; (3) the function is rebuilt in a restricted globals namespace exposing only \texttt{re}, \texttt{math}, and the tag constants; and (4) a probe call with dummy inputs verifies that the function returns a \texttt{list[float]} of the correct length. Rewards that fail any stage are discarded. Valid rewards are screened through 500-step GRPO training runs with the three base rewards always included alongside the candidate. The entire search takes approximately 40 hours on a single NVIDIA RTX 5090 GPU.

\subsection{Ensemble Reward Construction}

After the search is complete and all 50 individual rewards have been ranked, top-performing rewards are combined into ensembles. Each ensemble always retains the three base rewards alongside $K$ discovered rewards and sums all scores element-wise:
\begin{equation}
R_{\text{ensemble}}(o) = \sum_{b=1}^{3} R^{\text{base}}_b(o) + \sum_{k=1}^{K} R_k(o)
\label{eq:ensemble}
\end{equation}
In the configurations reported below, the count of rewards (e.g., ``top~3'' or ``top~5'') refers to the number of discovered rewards added on top of the three fixed base rewards. Seven ensemble configurations were evaluated. These configurations vary in size (3 to 10 discovered rewards) and selection strategy (ranking-based vs.\ diversity-based). The diversity-focused ensemble, \texttt{ens\_diverse\_rounds}, selects one discovered reward from each of the five rounds to maximize coverage of different reward strategies.

Algorithm~\ref{alg:framework} summarizes the complete framework, including the four-stage sandbox validation used to screen every generated reward function before it enters GRPO training.

\begin{algorithm}[t]
\caption{Search-Driven Reward Function Synthesis}
\label{alg:framework}
\begin{algorithmic}[1]
\Require Base policy $\pi_\theta$ (Llama-3.2-3B + LoRA); dataset $\mathcal{D}_{\text{train}}, \mathcal{D}_{\text{test}}$ (GSM8K); frontier LLM generator $\mathcal{G}$ (Kimi~K2); base rewards $\mathcal{B} = \{R^{\text{base}}_1, R^{\text{base}}_2, R^{\text{base}}_3\}$; rounds $N{=}5$; rewards per round $M{=}10$; GRPO steps $T{=}500$
\Ensure Ranked rewards $\mathcal{P}$; best ensemble $E^\star$
\State $\mathcal{P} \gets \emptyset$;\ $H \gets \emptyset$ \Comment{pool and history}
\For{$r = 1$ \textbf{to} $N$}
 \State $\texttt{prompt}_r \gets \textsc{BuildPrompt}(H, \text{samples from } \mathcal{D}_{\text{train}})$
 \State $\mathcal{C}_r \gets \mathcal{G}(\texttt{prompt}_r)$ \Comment{$M$ candidate reward sources}
 \ForAll{$R \in \mathcal{C}_r$}
 \State $s_1 \gets \textsc{CheckModules}(R)$ \Comment{Stage 1: imports}
 \State $s_2 \gets \textsc{BuildLockedDownFn}(R)$ \Comment{Stage 2: AST sandbox}
 \State $s_3 \gets \textsc{BuildSafeCallable}(R)$ \Comment{Stage 3: wrap \& resource limits}
 \State $s_4 \gets \textsc{ProbeCallable}(R)$ \Comment{Stage 4: dry-run on batch}
 \If{\textbf{not} $(s_1 \wedge s_2 \wedge s_3 \wedge s_4)$}
 \State \textbf{continue} \Comment{reject unsafe/invalid reward}
 \EndIf
 \State $\mathcal{R}_{\text{train}} \gets \mathcal{B} \cup \{R\}$
 \State $\pi_R \gets \textsc{GRPO}(\pi_\theta, \mathcal{D}_{\text{train}}, \mathcal{R}_{\text{train}}, T)$
 \State $(\text{acc}, P, \text{rec}, F_1) \gets \textsc{Evaluate}(\pi_R, \mathcal{D}_{\text{test}})$
 \State $\mathcal{P} \gets \mathcal{P} \cup \{(R, F_1, r)\}$
 \EndFor
 \State $H \gets \textsc{RankSummary}(\mathcal{P})$ \Comment{selection pressure}
\EndFor
\Statex
\State \textbf{Ensemble stage:}
\State $\mathcal{E} \gets \textsc{BuildEnsembleConfigs}(\mathcal{P})$ \Comment{7 configs, ranking \& diversity}
\ForAll{$E \in \mathcal{E}$ with discovered rewards $\{R_k\}_{k=1}^{K}$}
 \State $R_E(o) \gets \sum_{b=1}^{3} R^{\text{base}}_b(o) + \sum_{k=1}^{K} R_k(o)$
 \State $\pi_E \gets \textsc{GRPO}(\pi_\theta, \mathcal{D}_{\text{train}}, \{R_E\}, T)$
 \State $F_1^E \gets \textsc{Evaluate}(\pi_E, \mathcal{D}_{\text{test}})$
\EndFor
\State $E^\star \gets \arg\max_{E \in \mathcal{E}} F_1^E$
\State \Return $\mathcal{P},\ E^\star$
\end{algorithmic}
\end{algorithm}

\subsection{Training Configuration}
\label{sec:training}

GRPO training uses a group size of $G = 4$. The effective batch size is 4 (per-device batch size 1, gradient accumulation 4). The learning rate is $5 \times 10^{-6}$ with cosine annealing, warmup ratio 0.1, and weight decay 0.1. The optimizer is AdamW with 8-bit quantization. Maximum gradient norm is 1.0. Individual screening trials run for 500 steps, and ensemble training also uses 500 steps (except an initial pilot at 300 steps). The 500-step screening budget follows the recommendation in the official Unsloth GRPO documentation~\cite{unslothdocs}, which suggests training for at least 300 steps to obtain a reliable reward signal; we exceed this threshold by 67\% to balance ranking reliability against the wall-clock cost of evaluating 50 candidate rewards. The training pipeline is built on the TRL library~\cite{vonwerra2020trl} with Unsloth~\cite{unsloth} for efficient 4-bit LoRA fine-tuning.

\subsection{Evaluation}
\label{sec:metrics}

All evaluations use greedy decoding (temperature = 0). The model output is parsed to extract the content between \texttt{<solution>} tags. Four metrics are computed over the 1,319-question test set: precision (TP / (TP + FP)), recall (TP / (TP + FN)), F1 (harmonic mean of precision and recall), and accuracy (the fraction of the 1,319 test questions for which the policy produces an extracted answer that matches the ground truth). In this setting TP counts questions whose extracted answer matches the ground truth, FP counts questions whose extracted answer does not match, and FN counts questions that produce no parseable \texttt{<solution>} answer. Outputs that fail strict-format extraction are therefore counted as wrong rather than skipped, so the reported accuracy is the standard exact-match rate under strict tag-format parsing. To isolate the contribution of strict format parsing from reasoning quality, supplementary Section~S6 additionally reports exact-match accuracy under flexible parsing, in which a numeric-token regex over the full completion is used as a fallback when strict extraction fails. F1 is the primary ranking metric because it balances answer correctness against format compliance. Bootstrap 95\% confidence intervals (10,000 resamples with replacement over the 1,319 test items) are reported for all headline accuracies, and pairwise McNemar tests are reported for the seven ensemble configurations. Because 21 pairwise tests are performed, the corresponding Bonferroni-corrected significance threshold is $\alpha\,=\,0.05/21\,\approx\,0.0024$.

Five off-the-shelf models of comparable size (2B--4B parameters) are evaluated as baselines on the same test set under identical conditions: Qwen~3/3.5 (4B, 2B) and Phi-4-Mini (3.8B, reasoning variant).

The evaluation protocol is designed so that the contribution of the discovered rewards can be read off by comparison rather than asserted in isolation. Two controls are evaluated under identical decoding, parsing, and bootstrap procedures: a base-rewards-only configuration, which isolates the effect of adding any discovered reward on top of the three protected base rewards, and a random-5 configuration, which isolates the effect of ranked selection from the additive effect of merely combining five rewards. The headline ensemble is additionally retrained under three independent GRPO seeds (Section~\ref{sec:seeds}) to characterize seed-induced variance. Because all configurations --- discovered, base-only, random, and multi-seed --- are evaluated under the same protocol, the comparisons reported in Section~\ref{sec:results} are internal to the pipeline and the relative ordering across configurations is the primary object of inference.

\section{Results}
\label{sec:results}

\subsection{Search Trajectory}

The search over five rounds produced 50 candidate reward functions. The F1 scores of all 50 rewards (sorted from highest to lowest, supplementary Fig.~S1) reveal a steep drop-off: the top 6 rewards achieve F1 scores above 0.75, while the majority cluster between 0.55 and 0.65. The worst reward achieves an F1 of only 0.059. Only approximately 12\% of rewards exceed 0.60 accuracy.

The top five individual rewards are: \texttt{thinking\_steps\_count} (F1\,=\,0.787, Round~5), \texttt{thinking\_has\_calc} (F1\,=\,0.786, Round~5), \texttt{numeric\_reasoning\_depth} (F1\,=\,0.783, Round~4), \texttt{step\_by\_step\_accuracy} (F1\,=\,0.779, Round~2), and \texttt{reasoning\_completeness} (F1\,=\,0.754, Round~2). Three of these five focus on the quality and structure of the chain-of-thought reasoning; the other two evaluate completeness and step-by-step correctness. We caution that these are unverified structural rewards and could in principle be exploitable by a policy that only pads its reasoning block; we audit this concern explicitly in Section~\ref{sec:hacking}. For the same reason these rewards are never used alone: they are summed with the three correctness-checking base rewards. The full F1 ranking of all 50 rewards is given in the supplementary material (Fig.~S1).

\subsubsection{Performance Variation Over Rounds}

\begin{figure}[!t]
\centering
\includegraphics[width=\columnwidth]{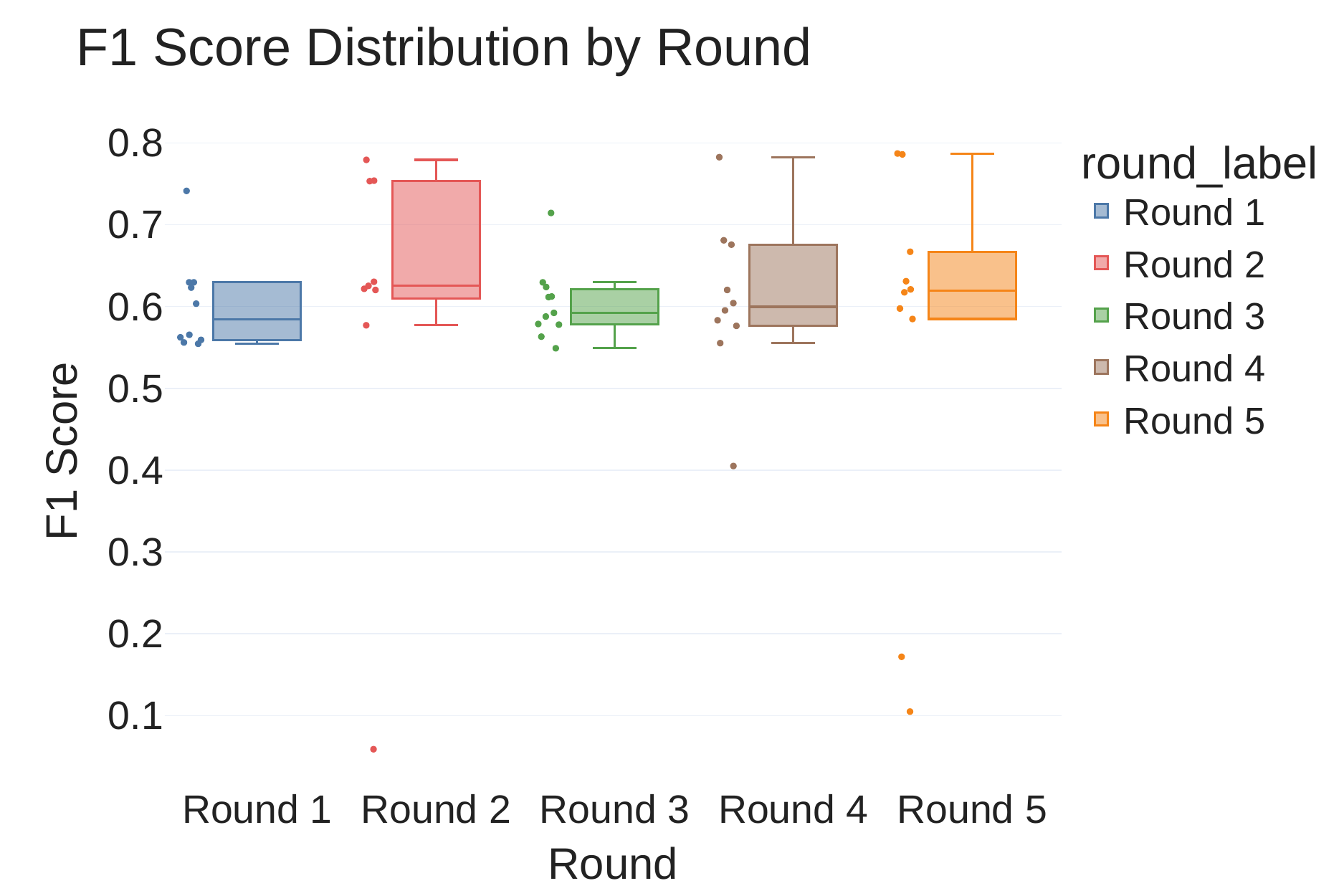}
\caption{Box plot of F1 scores by round. The median F1 increases from Round~1 through Round~5, though with a dip in Round~3.}
\label{fig:box_f1_round}
\end{figure}

Fig.~\ref{fig:box_f1_round} presents box plots of F1 scores grouped by round. The mean F1 increases from 0.596 in Round~1 to 0.632 in Round~5. The improvement is not strictly monotonic: Round~3 dips to 0.604 before recovery in Round~4 (0.618) and Round~5. This non-monotonic pattern may reflect the search exploring a different region of reward function space that did not yield strong rewards on average, even as the system continued accumulating information about effective strategies.

\begin{figure}[!t]
\centering
\includegraphics[width=\columnwidth]{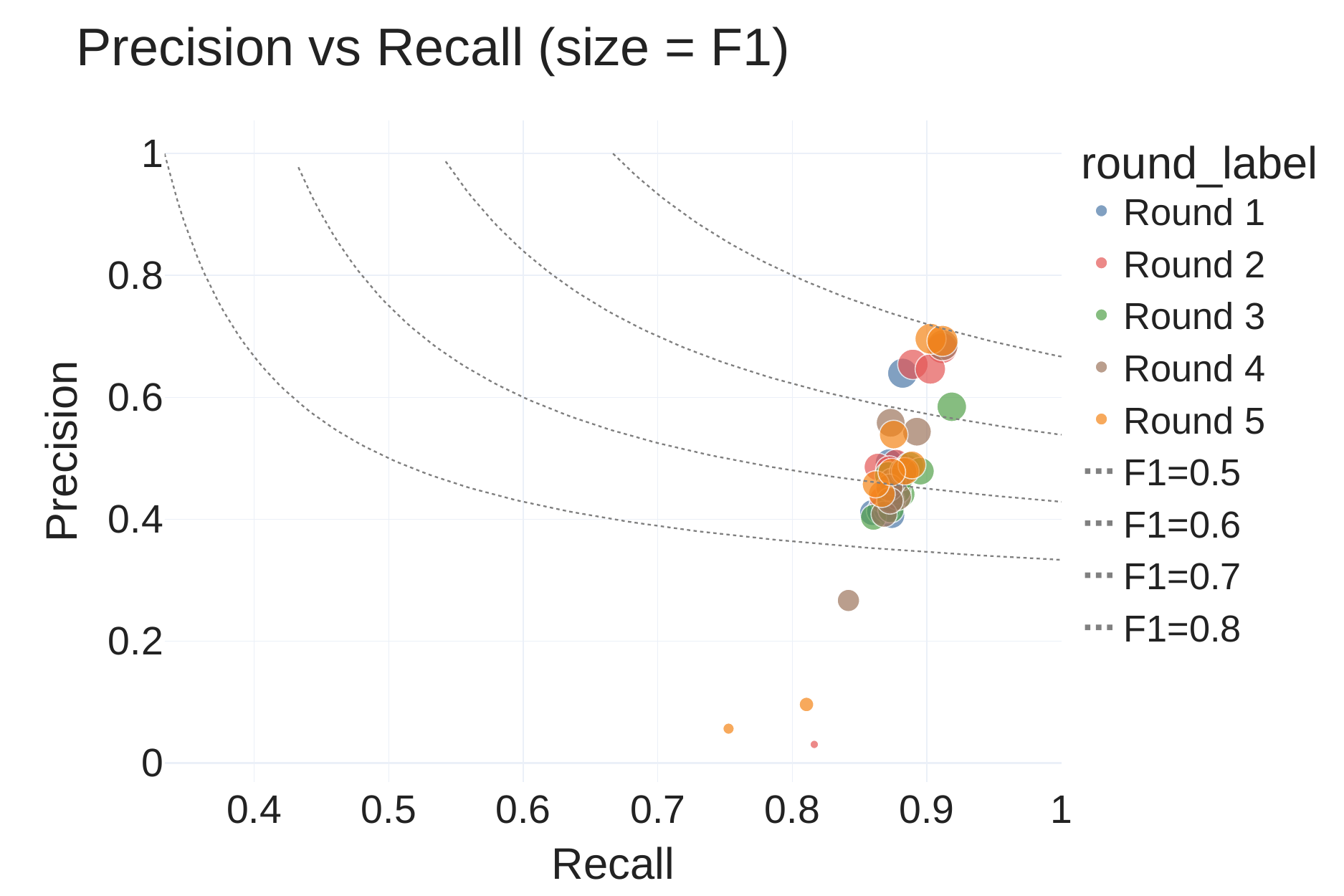}
\caption{Precision versus recall for all 50 individual reward functions, colored by round.}
\label{fig:precision_recall}
\end{figure}

A precision--recall scatter over all 50 rewards (Fig.~\ref{fig:precision_recall}) shows that most rewards cluster where recall is high (above 0.8) but precision is moderate (0.5--0.7), indicating they encourage the model to always produce a formatted answer without strongly discriminating correct from incorrect; the top-performing rewards achieve a better balance with both metrics above 0.7.

\subsection{Analysis of Top-Performing Reward Functions}

\subsubsection{Reward Function Code}

Listing~\ref{lst:thinking_steps_count} shows the top-ranked reward function. It counts distinct non-empty lines in the \texttt{<thinking>} block and assigns graduated scores of +3, +2, or +1 for three, two, or one reasoning steps respectively, and $-1$ for a missing thinking block. This reward does not check answer correctness; the three base rewards provide that signal.

\begin{lstlisting}[caption={Top-ranked reward: \texttt{thinking\_steps\_count} (F1\,=\,0.787, Round~5). Rewards multi-line reasoning structure without checking correctness.}, label={lst:thinking_steps_count}, float=t]
def thinking_steps_count(prompts, completions,
 answer, **kwargs):
 import re
 scores = []
 for completion, _ in zip(completions, answer):
 resp = completion[0]['content']
 m = re.search(
 r'<thinking>(.*?)</thinking>',
 resp, re.DOTALL)
 if not m:
 scores.append(-1.0)
 continue
 steps = [s for s in
 re.split(r'\n+', m.group(1).strip())
 if s.strip()]
 if len(steps) >= 3: scores.append(3.0)
 elif len(steps) == 2: scores.append(2.0)
 elif len(steps) == 1: scores.append(1.0)
 else: scores.append(-1.0)
 return scores
\end{lstlisting}

The second-ranked reward, \texttt{thinking\_has\_calc} (F1\,=\,0.786, Round~5), applies a binary check for whether the thinking section contains arithmetic operations using the regex pattern \texttt{[0-9]+\textbackslash{}s*[+\textbackslash{}-*/]}. A score of +2 is awarded when explicit calculations are present and 0 otherwise, with $-1$ for a missing thinking block. The highest-ranked hybrid reward, \texttt{step\_by\_step\_accuracy} (F1\,=\,0.779, Round~2), combines arithmetic-step detection with answer correctness and assigns graduated scores from $+4$ to $-1$; its full source is given in the supplementary material (Listing~S2).

The top ten discovered rewards group into three categories: pure-process (4 rewards, mean F1 = 0.759), hybrid process plus outcome (4 rewards, mean F1 = 0.757), and pure-outcome (2 rewards, mean F1 = 0.695). The pure-process rewards achieve the highest mean F1 because the three base rewards already provide strong correctness signals; adding redundant outcome verification yields diminishing returns, while process-oriented signals such as step counts, arithmetic detection, and numeric density are complementary. The full taxonomy table is given in the supplementary material (Table~S1).

\subsubsection{Empirical Patterns and Their Limits}

Three patterns appear in the top ten rewards. The top three discovered rewards (\texttt{thinking\_steps\_count}, \texttt{thinking\_has\_calc}, \texttt{numeric\_reasoning\_depth}) all evaluate reasoning structure rather than re-checking the final answer, with mean F1 0.785 versus 0.695 for the two pure outcome rewards in the top ten; we frame this complementarity as suggestive given the small sample (3 vs.\ 2). Eight of the top ten use three or more distinct score levels rather than binary $\{0,1\}$, consistent with the known benefits of graduated reward shaping for credit assignment. Code length is not predictive: over the full pool of 50 the correlation between line count and F1 is $r\,{=}\,0.04$, and the bottom-ten mean line count (20.5) is comparable to the top ten (24.0).

Three discovered rewards in the pool of 50 yielded catastrophic GRPO performance (F1 well below the 0.609 base-only baseline): \texttt{efficiency\_vs\_accuracy} (F1\,=\,0.059), \texttt{thinking\_no\_answer\_leak} (F1\,=\,0.105), and \texttt{thinking\_length\_range} (F1\,=\,0.172). The common thread is that aggressive penalty terms with no correctness gate dominate the gradient before the policy has discovered any high-reward region of behavior, a known failure mode of reward shaping in sparse-reward RL~\cite{ng1999policy}. This is one practical reason to keep the three base rewards on at all times. A full per-reward post-mortem is given in the supplementary material (Section~S3).

\subsection{Ensemble Results}

Table~\ref{tab:ensemble_results} summarizes the performance of all ensemble configurations together with bootstrap 95\% confidence intervals. The numerically best ensemble is \texttt{ens\_diverse\_rounds}, which combines the three base rewards with five discovered rewards drawn one per round. It achieves F1\,=\,0.795 and accuracy\,=\,0.660 (95\% CI [0.635,\,0.686]). However, the bootstrap intervals of the five largest ensembles (5--10 discovered rewards) all overlap substantially. Two control rows sit below the main results: ``Base only (3 rewards)'' is GRPO with only the three protected base rewards under identical hyperparameters and seed and isolates the contribution of the discovered rewards; ``ens\_random5'' replaces ranked selection with five rewards drawn uniformly at random from the pool of 50 and isolates the contribution of selection from that of merely adding five rewards. The random draw included \texttt{efficiency\_vs\_accuracy}, one of the catastrophic rewards (F1\,=\,0.059 in isolation), which collapses training to F1\,=\,0.047 even with the three base rewards present.

\begin{table}[t]
\centering
\caption{Ensemble performance on GSM8K (1,319 questions). ``\#R'' counts discovered rewards added to the three fixed base rewards. ``Accuracy'' is exact-match accuracy under strict tag-format parsing, defined in Section~\ref{sec:metrics}; 95\% bootstrap CI in brackets. The five large ensembles ($\geq 5$ discovered rewards) are statistically indistinguishable under McNemar with Bonferroni correction (supplementary Table~S2).}
\label{tab:ensemble_results}
\setlength{\tabcolsep}{3pt}
\begin{tabular}{lcccc}
\toprule
Configuration & \# R & F1 & Accuracy\ [95\% CI] & Rec. \\
\midrule
ens\_diverse\_rounds & 5 & \textbf{0.795} & \textbf{0.660} [0.635, 0.686] & 0.892 \\
ens\_top5\_500 & 5 & 0.789 & 0.651 [0.626, 0.677] & 0.909 \\
ens\_top10 & 10 & 0.787 & 0.648 [0.622, 0.673] & \textbf{0.911} \\
ens\_correctness & 5 & 0.784 & 0.644 [0.619, 0.670] & 0.883 \\
ens\_top7 & 7 & 0.782 & 0.641 [0.616, 0.667] & 0.914 \\
ens\_top3 & 3 & 0.738 & 0.585 [0.558, 0.611] & 0.759 \\
final\_ens\_300 & 5 & 0.743 & 0.591 [0.564, 0.617] & 0.896 \\
\midrule
Best Indiv. & 1 & 0.787 & 0.649 [0.624, 0.674] &; \\
\midrule
ens\_random5 & 5 & 0.047 & 0.024 [0.017, 0.033] & 0.762 \\
Base only (3 rewards) & 0 & 0.609 & 0.438 [0.412, 0.466] &; \\
\bottomrule
\end{tabular}
\end{table}

\subsubsection{Statistical Significance}

Pairwise McNemar tests over the per-question correctness vectors of the seven ensembles produce 21 comparisons, requiring a Bonferroni-corrected significance threshold of $\alpha\,=\,0.0024$. The five large ensembles ($\geq 5$ discovered rewards trained for 500 steps) are not statistically distinguishable from one another at this threshold ($p$-values between 0.04 and 0.39 for all pairs). Both the three-reward configuration and the five-reward 300-step pilot are significantly worse than every five-reward-or-larger ensemble ($p < 10^{-6}$ for the relevant pairs). The full $p$-value table is provided in the supplementary material (Table~S2).

\subsubsection{Robustness Over Seeds}
\label{sec:seeds}

To probe whether the reported ranking is sensitive to the GRPO training seed, the \texttt{ens\_diverse\_rounds} configuration was retrained from scratch under two additional seeds (1234 and 2025) on the same fixed Llama-3.2-3B base, with identical hyperparameters and an unchanged 1{,}319-question evaluation set. Across the three seeds, F1 is $0.785 \pm 0.013$ (range 0.771 to 0.795) and accuracy is $0.646 \pm 0.017$. Two observations follow. First, the diversity ensemble's lead over the top-5 ranking ensemble (default-seed gap of $0.795 - 0.789 = 0.006$) is well inside the seed-to-seed variation, and one of the three seeds (2025, F1\,=\,0.771) trails the default-seed top-5 ensemble. The diversity-vs-ranking comparison is therefore reported as exploratory rather than confirmatory throughout the paper. Second, even the worst seed of the diversity ensemble (F1\,=\,0.771) substantially exceeds both the three-reward configuration (F1\,=\,0.738) and the base-rewards-only baseline (F1\,=\,0.609) by margins much larger than the seed-to-seed spread. The qualitative ordering ``base only $<$ three-reward $<$ five large ensembles'' is therefore robust to seed choice, while finer-grained ranking among the five large ensembles is not. Per-seed numbers are tabulated in the supplementary material (Table~S3).

\begin{figure}[!t]
\centering
\includegraphics[width=\columnwidth]{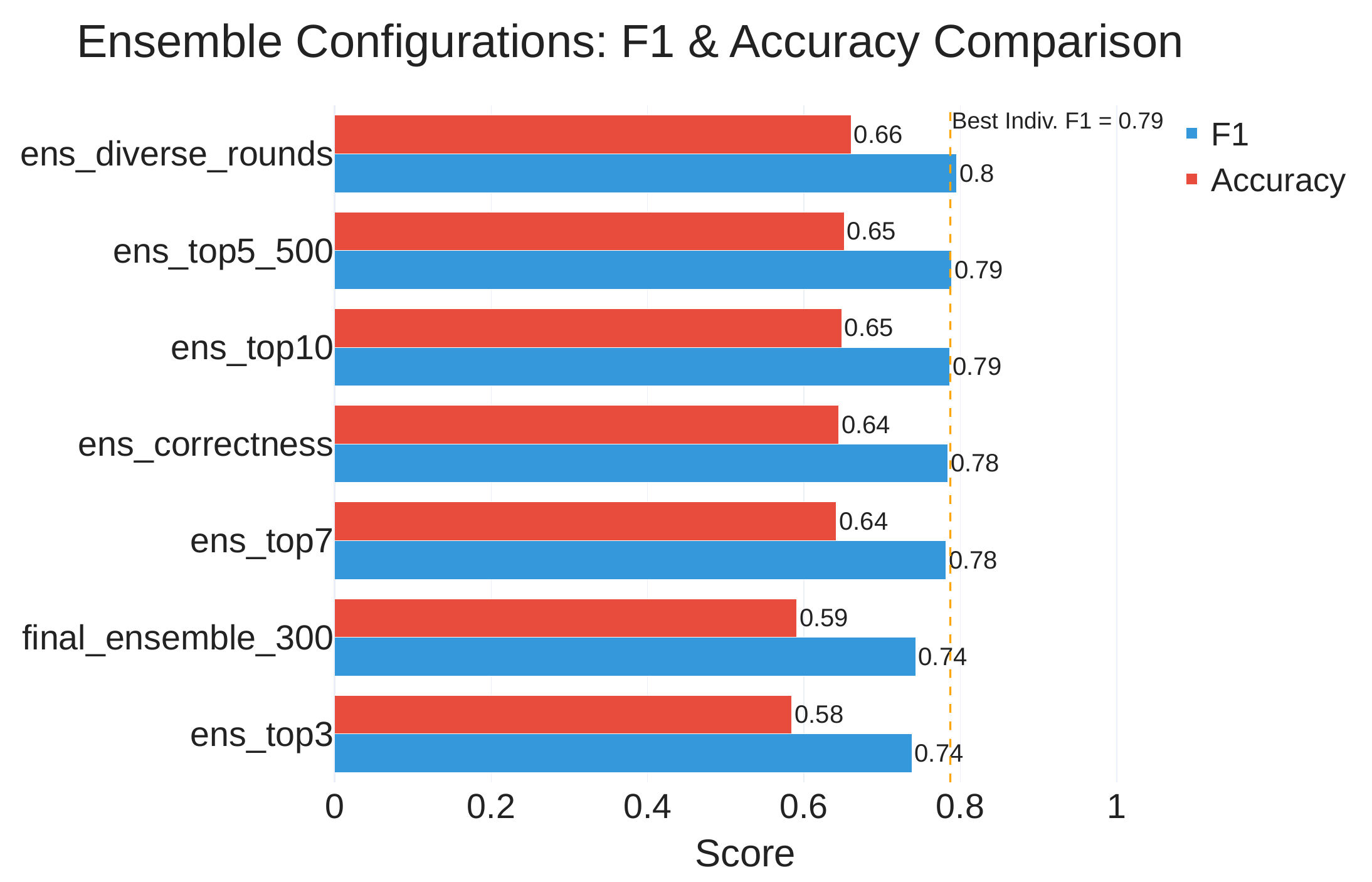}
\caption{All ensemble configurations compared on four evaluation metrics. The five large ensembles cluster tightly; only the three-reward and 300-step configurations fall out of the band.}
\label{fig:ensemble_comparison}
\end{figure}

A visualization of all ensemble configurations on the four evaluation metrics is shown in Fig.~\ref{fig:ensemble_comparison}.

\subsection{Reward-Hacking and Format-Penalty Audits}
\label{sec:hacking}

Three of the top five discovered rewards count structural features of the \texttt{<thinking>} block without verifying answer correctness, with \texttt{thinking\_steps\_count} (F1\,=\,0.787) the most aggressive: it pays $+3.0$ for at least three newline-separated lines and is the textbook target for length-padding hacks. On the 1{,}319 GSM8K test items the trained policy emits a mean of 6.13 lines inside \texttt{<thinking>} (median 6), so the $\geq 3$ threshold is exceeded organically rather than minimally. Splitting outputs by correctness, correct responses contain 6.47 lines on average and incorrect ones 5.51, with arithmetic-step-pattern density 0.63 vs.\ 0.51 respectively. If the model were padding empty lines, the correct vs.\ incorrect distributions would be similar or inverted; instead, correct answers are accompanied by both more lines and a higher density of arithmetic content. The simplest line-padding hypothesis is therefore not supported by the per-question evidence.

A separate concern is the strict tag-format penalty in the accuracy column. Re-evaluating the trained policies under flexible answer extraction (numeric-token regex over the full completion as fallback) shifts accuracy by +0.012 to +0.020 for the five large ensembles and by +0.092 for the three-reward ensemble. The format penalty for the large ensembles is therefore small; the three-reward configuration is producing correct reasoning that frequently lands in a non-canonical solution tag. Full per-configuration numbers and a longer audit discussion are provided in the supplementary material (Sections~S6 and~S7).

\subsection{Baseline Comparison}

Table~\ref{tab:baseline_results} presents the performance of five off-the-shelf models in the 2--4B parameter band. The trained Llama-3.2-3B model achieves the highest F1 (0.795) and accuracy (0.660) among models in this size range, surpassing all evaluated baselines. The remaining baselines span an accuracy range from 0.403 to 0.634.

\begin{table}[t]
\centering
\caption{Off-the-shelf model performance on GSM8K vs.\ GRPO-trained model. ``Accuracy'' for the GRPO-trained model is the strict tag-format exact-match accuracy of Section~\ref{sec:metrics}; for the off-the-shelf baselines it equals their native (already flexible) numeric-extraction accuracy. ``Acc.\ (flex)'' for the GRPO model uses flexible parsing (Section~\ref{sec:hacking}; full table in supplementary Section~S6).}
\label{tab:baseline_results}
\setlength{\tabcolsep}{3pt}
\begin{tabular}{lcccc}
\toprule
Model & Params & Accuracy & Acc.\ (flex) & F1 \\
\midrule
qwen3:4b & 4B & 0.634 &; & 0.776 \\
phi4-mini:3.8b & 3.8B & 0.629 &; & 0.773 \\
qwen3.5:4b & 4B & 0.449 &; & 0.620 \\
phi4-mini-reas. & 3.8B & 0.409 &; & 0.581 \\
qwen3.5:2b & 2B & 0.403 &; & 0.575 \\
\midrule
GRPO (ours) & 3B & \textbf{0.660} & \textbf{0.673} & \textbf{0.795} \\
\bottomrule
\end{tabular}
\end{table}

A direct read of Table~\ref{tab:baseline_results} is that the GRPO-trained Llama-3.2-3B leads its size band on both strict and flexible accuracy and on F1.

The contribution of this paper is methodological rather than a claim to a new state of the art on GSM8K. The question is whether the reward specification used during RL post-training can be treated as a reliably searchable object. The Llama-3.2-3B base was chosen because a full search over 50 rewards plus seven ensembles plus three reproducibility seeds fits inside an academic single-GPU budget; the same protocol applies unchanged to any larger base, and wall-clock cost grows roughly linearly with model size. The evidence that matters is internal to the pipeline: the F1 of generated rewards rises from round to round (Fig.~\ref{fig:box_f1_round}); the random-5 control collapses to F1\,=\,0.047 while ranked ensembles cluster near 0.79; the qualitative ordering ``base only $<$ three-reward $<$ five large ensembles'' holds over three independent training seeds. Re-running the pipeline on a stronger base to test whether the discovered rewards stack on top of a higher starting point is a natural item of future work (Section~\ref{sec:future-work}).

\subsection{Computational Cost}

The full search over 5 rounds and 50 rewards takes approximately 40 hours of wall-clock time on a single NVIDIA RTX 5090 GPU (32~GB VRAM); per-trial timing details are reported in supplementary Section~S8.

\section{Discussion}

The results support three main conclusions about search-driven reward optimization for LLM reasoning.

First, iterative LLM-guided generation conditioned on ranked feedback yields measurable improvement over rounds. The mean F1 of generated rewards increased from 0.596 in Round~1 to 0.632 in Round~5, and the three highest-scoring individual rewards all emerged from the later rounds. The non-monotonic dip in Round~3 suggests that the search occasionally explores qualitatively different regions of reward space, which is desirable to prevent premature convergence on a single strategy.

Second, combining several discovered rewards into ensembles produces stronger training signals than the best single reward. All five large ensembles (5 to 10 discovered rewards added to the three base rewards, 500 steps) cluster between F1 0.78 and 0.80 and accuracy 0.64 to 0.66, and pairwise McNemar tests with Bonferroni correction show that they are not statistically distinguishable from one another. The three-reward configuration and the 300-step pilot fall significantly below this band. Two additional pieces of evidence sharpen this picture. A three-seed re-training of the best ensemble yields F1 of $0.785 \pm 0.013$ (Section~\ref{sec:seeds}); the original 0.795 vs.\ 0.789 gap therefore sits inside seed-to-seed variation. A random-5 control drawn from the same pool collapses to F1\,=\,0.047, far below the base-only baseline; the ranked feedback loop, not the mere act of stacking five rewards, is what produces the gain.

Third, the discovered rewards are largely structural and could in principle be hackable, but the audit in Section~\ref{sec:hacking} shows that the most exploitable reward (\texttt{thinking\_steps\_count}) does not, in practice, induce empty line padding under this base model and budget; correct answers carry both more lines and a higher density of arithmetic content than incorrect ones. Combined with the format-penalty analysis in supplementary Section~S6, which shows that strict tagging accounts for only $\sim 1$--$2\%$ of the absolute accuracy of the large ensembles, the discovered rewards behave as expected complements to the three base correctness rewards rather than as independent gameable surrogates.

\subsection{Limitations}

A number of design choices in this work define a clear scope for the present study. The focused setup, in which every component of the pipeline other than the reward specification is held fixed, makes the results reproducible at the scale of an academic compute budget~\cite{khandelwal2024100} and isolates the effect of changing the reward specification on downstream reasoning performance. Mapping how the discovered rewards behave when the benchmark, base model, generator, or optimizer changes is a natural extension that we leave to follow-up studies (Section~\ref{sec:future-work}).

The screening budget of 500 GRPO steps per candidate exceeds the 300-step minimum recommended by the Unsloth GRPO documentation~\cite{unslothdocs} and is appropriate for ranking 50 candidates on a single GPU, but rewards whose effects only emerge after substantially longer training cannot be separated at this budget; re-training the top-ranked rewards for 2,000 steps is the cleanest way to test how stable the 500-step ranking is asymptotically. Multi-seed evidence is currently reported for the diversity-selected ensemble only (Section~\ref{sec:seeds}), which suffices to show that the qualitative ordering ``base only $<$ three-reward $<$ five large ensembles'' is robust to seed but that finer-grained ranking among the five large ensembles is not. The ensemble mechanism is also intentionally simple: rewards are summed with equal weights, and learned weighting is left as a complementary direction.

A more structural limitation concerns the transferability of the discovered reward code itself. The 50 reward functions are written against the \texttt{<thinking>}/\texttt{<solution>} tag schema and GSM8K-style numeric final answers; evaluating them on a different schema or on benchmarks with non-numeric answers requires rewriting the regex and answer-comparison logic.

\section{Conclusion}

This paper investigated whether the reward function used in reinforcement learning for language-model reasoning can be treated as an object of search rather than a fixed design choice. A search-driven reward synthesis framework was developed and evaluated on the GSM8K mathematical reasoning benchmark over five rounds. The search produced 50 candidate rewards and seven ensemble configurations.

The search yielded measurable improvement over rounds, with the top-performing rewards emerging from later rounds. Combining several discovered rewards into ensembles produced stronger training signals than any single reward; under multiple-comparison correction, the five large ensembles are statistically indistinguishable from one another, while the three-reward and 300-step configurations are significantly worse. The diversity-vs-ranking comparison is therefore exploratory rather than confirmatory. A reward-hacking audit on the most exploitable structural reward shows that, in practice, correct responses contain more reasoning steps and a higher density of arithmetic operations than incorrect ones, which is the opposite of the line-padding failure mode predicted by a naive reading of the reward.

\subsection{Future Work}
\label{sec:future-work}

Future directions include applying the same pipeline to other base models of similar or larger scale (e.g., the Qwen-3, Phi-4, Mistral, Llama-3, and Gemma~3 families spanning 2B to 8B parameters) to test whether the discovered rewards stack on top of different starting points; comparing alternative reward generators (GPT, Claude, Gemini, DeepSeek) against Kimi K2 to characterize how generator choice influences which rewards are discovered; comparing alternative policy optimizers (PPO, RLOO, DPO) against GRPO; population-based strategies with explicit crossover, mutation, and diversity maintenance; multi-seed runs (3--5 seeds per configuration) on every ensemble; longer-horizon screening (e.g., 2{,}000 GRPO steps for the top five rewards); held-out reasoning benchmarks beyond GSM8K (e.g., MATH, AIME, MMLU-STEM, BBH); and learned ensemble weighting in place of equal-weight summation.

\subsection*{Reproducibility}

All code, generated reward functions, and per-question correctness vectors used to compute the bootstrap CIs and McNemar tests are released at \url{https://github.com/INQUIRELAB/search-reward-rl}. The release sandboxes external API keys behind environment variables, ships the four-stage validation pipeline, and includes scripts to regenerate every table and figure in this paper. Extended results (full reward-taxonomy table, McNemar p-value matrix, three-seed table, strict-vs-flexible-parse comparison, long-form hacking audit, full F1 ranking, full source of the top hybrid reward, and per-trial timing) are in the supplementary material.

\onecolumn
\renewcommand{\thesection}{S\arabic{section}}
\renewcommand{\thetable}{S\arabic{table}}
\renewcommand{\thefigure}{S\arabic{figure}}
\renewcommand{\thelstlisting}{S\arabic{lstlisting}}
\renewcommand{\thealgorithm}{S\arabic{algorithm}}
\renewcommand{\theHsection}{supp.\arabic{section}}
\renewcommand{\theHtable}{supp.\arabic{table}}
\renewcommand{\theHfigure}{supp.\arabic{figure}}
\renewcommand{\theHlstlisting}{supp.\arabic{lstlisting}}
\renewcommand{\theHalgorithm}{supp.\arabic{algorithm}}
\setcounter{section}{0}
\setcounter{table}{0}
\setcounter{figure}{0}
\setcounter{lstlisting}{0}
\setcounter{algorithm}{0}

\section{Reward-Function Code Listings}
\label{sec:supp-listings}

This appendix gives the full Python source of two top-ranked discovered rewards. Listing~\ref{lst:supp-thinking-steps} is the highest-F1 reward and is referenced from the body of the paper. Listing~\ref{lst:supp-step-accuracy} is the highest-ranked hybrid reward (process plus outcome).

\begin{figure}[h]
\captionof{lstlisting}{Top-ranked reward \texttt{thinking\_steps\_count} (F1 = 0.787, Round 5).}
\label{lst:supp-thinking-steps}
\vspace{0.6em}
\begin{lstlisting}
def thinking_steps_count(prompts, completions, answer, **kwargs):
    import re
    scores = []
    for completion, _ in zip(completions, answer):
        resp = completion[0]['content']
        m = re.search(r'<thinking>(.*?)</thinking>', resp, re.DOTALL)
        if not m:
            scores.append(-1.0)
            continue
        steps = [s for s in re.split(r'\n+', m.group(1).strip()) if s.strip()]
        if len(steps) >= 3: scores.append(3.0)
        elif len(steps) == 2: scores.append(2.0)
        elif len(steps) == 1: scores.append(1.0)
        else: scores.append(-1.0)
    return scores
\end{lstlisting}
\end{figure}

\begin{figure}[h]
\captionof{lstlisting}{Highest-ranked hybrid reward \texttt{step\_by\_step\_accuracy} (F1 = 0.779, Round 2). Detects calculation expressions in the thinking block and combines them with answer-correctness gating.}
\label{lst:supp-step-accuracy}
\vspace{0.6em}
\begin{lstlisting}
def step_by_step_accuracy(prompts, completions, answer, **kwargs):
    import re
    scores = []
    for completion, true_answer in zip(completions, answer):
        resp = completion[0]['content']
        thinking = re.search(r'<thinking>(.*?)</thinking>', resp, re.DOTALL)
        solution = re.search(r'<solution>(.*?)</solution>', resp, re.DOTALL)
        if not thinking or not solution:
            scores.append(-1.0)
            continue
        think_text = thinking.group(1).strip()
        sol_text = solution.group(1).strip()
        steps = re.findall(r'(\d+(?:\.\d+)?)\s*([+*\-/])\s*(\d+(?:\.\d+)?)', think_text)
        try:
            pred = float(sol_text.replace(',', ''))
            true_val = float(str(true_answer).replace(',', ''))
            correct = abs(pred - true_val) < 1e-6
        except (ValueError, TypeError):
            correct = False
        if correct and len(steps) >= 2: scores.append(4.0)
        elif correct and len(steps) >= 1: scores.append(3.0)
        elif correct: scores.append(2.0)
        elif len(steps) >= 2: scores.append(0.0)
        else: scores.append(-1.0)
    return scores
\end{lstlisting}
\end{figure}

\section{Reward Taxonomy of Top-Ten Discovered Rewards}
\label{sec:supp-taxonomy}

\begin{table}[h]
\centering
\caption{Taxonomy of the top 10 discovered reward functions grouped by evaluation strategy. Pure process rewards achieve the highest mean F1 when combined with the three base rewards.}
\label{tab:supp-reward_taxonomy}
\begin{tabular}{llcc}
\toprule
Category & Rewards & Count & Mean F1 \\
\midrule
Pure Process & thinking\_steps\_count, & 4 & 0.759 \\
 & thinking\_has\_calc, & & \\
 & numeric\_reasoning\_depth, & & \\
 & thinking\_covers\_steps & & \\
\addlinespace
Hybrid & step\_by\_step\_accuracy, & 4 & 0.757 \\
(Process+Outcome) & reasoning\_completeness, & & \\
 & thinking\_quality\_score, & & \\
 & thinking\_length\_bonus & & \\
\addlinespace
Pure Outcome & correctness\_with\_units, & 2 & 0.695 \\
 & solution\_length\_penalty & & \\
\bottomrule
\end{tabular}
\end{table}

The three patterns observed in the top ten rewards are: (i)~pure-process rewards complement the base correctness signal (mean F1 = 0.785 for the top three pure-process rewards vs.\ 0.695 for the top two pure-outcome rewards); (ii)~graduated scoring with three or more levels appears in 8 of 10 top rewards; (iii)~code length is not predictive of reward quality over the full pool ($r = 0.04$ between line count and F1 across all 50 rewards; the bottom-ten mean line count of 20.5 is comparable to the top-ten mean of 24.0).

\section{Failure Modes of Low-F1 Discovered Rewards}
\label{sec:supp-failures}

Three of the 50 discovered rewards yielded GRPO performance well below the base-rewards-only baseline (F1 = 0.609). Their failure modes reveal characteristic pitfalls in reward shaping for sparse-reward RL.

\textbf{\texttt{efficiency\_vs\_accuracy} (F1 = 0.059, accuracy = 0.030).} This reward grants $+5$ for correct answers with $\leq 30$ words of thinking, with smaller bonuses for longer correct thinking. The extraction rate at evaluation is 0.993 but 1{,}270 of 1{,}319 final answers are wrong. The pathology is that a strong bonus on short thinking dominates the gradient before the model has learned to solve any problems. The policy collapses onto producing a near-empty \texttt{<thinking>} block followed by an arbitrary number; correct-answer gating becomes irrelevant because correct answers almost never occur during the brittle initial regime.

\textbf{\texttt{thinking\_no\_answer\_leak} (F1 = 0.105, accuracy = 0.055).} This reward returns $-1$ if the ground-truth answer string appears anywhere in \texttt{<thinking>} and $+1$ otherwise. It is intended to discourage the model from copying the answer into reasoning, but the GRPO advantage signal teaches the policy to never write the target value in \texttt{<thinking>}. The model learns to produce reasoning that systematically avoids the correct number and then guesses in \texttt{<solution>}, with predictably catastrophic results.

\textbf{\texttt{thinking\_length\_range} (F1 = 0.172, accuracy = 0.094).} This reward gives $+3$ for thinking length in $[100, 400]$ characters and $-1$ otherwise. The cliff at 400 characters truncates harder problems mid-derivation: any chain-of-thought that requires more than approximately 80 tokens of working is actively penalized, and the policy converges on artificially short, incomplete derivations.

The common thread across the three failures is that aggressive penalty terms with no correctness gate dominate the gradient before the policy has discovered any high-reward region of behavior. This is a known failure mode of reward shaping in sparse-reward RL~\cite{ng1999policy}, and it is one practical reason to keep the three base rewards on at all times: they provide a stable correctness signal that prevents purely structural penalties from collapsing the policy onto degenerate behavior. The random-5 control reported in the main paper (Section~III.E of the main text) draws \texttt{efficiency\_vs\_accuracy} in its random selection, which is the dominant cause of its collapse to F1 = 0.047.

\section{Pairwise McNemar Tests}
\label{sec:supp-mcnemar}

Pairwise McNemar tests over the per-question correctness vectors of the seven ensembles produce 21 comparisons. With Bonferroni correction the significance threshold is $\alpha = 0.05/21 = 0.0024$.

\begin{table}[h]
\centering
\caption{McNemar pairwise $p$-values for the seven ensemble configurations. With Bonferroni correction over 21 comparisons, $p < 0.0024$ is significant. ``n.s.'' denotes not significant after correction.}
\label{tab:supp-mcnemar}
\begin{tabular}{lc}
\toprule
Comparison & $p$ \\
\midrule
ens\_diverse\_rounds vs.\ ens\_top5\_500 & 0.39 (n.s.) \\
ens\_diverse\_rounds vs.\ ens\_top10 & 0.22 (n.s.) \\
ens\_diverse\_rounds vs.\ ens\_correctness & 0.14 (n.s.) \\
ens\_diverse\_rounds vs.\ ens\_top7 & 0.04 (n.s.\ after Bonf.) \\
ens\_diverse\_rounds vs.\ ens\_top3 & $< 10^{-6}$ \\
ens\_diverse\_rounds vs.\ final\_ens\_300 & $< 10^{-6}$ \\
ens\_top5\_500 vs.\ ens\_top3 & $< 10^{-6}$ \\
ens\_top3 vs.\ final\_ens\_300 & 0.68 (n.s.) \\
\bottomrule
\end{tabular}
\end{table}

The five large ensembles ($\geq 5$ discovered rewards trained for 500 steps) are not statistically distinguishable from each other under multiple-comparison correction. Both the three-reward configuration and the five-reward 300-step pilot are significantly worse than the five large ensembles.

\section{Multi-Seed Robustness of the Best Ensemble}
\label{sec:supp-seeds}

\begin{table}[h]
\centering
\caption{Three-seed reproducibility of \texttt{ens\_diverse\_rounds} on GSM8K (1{,}319 questions). All three seeds share the same five discovered rewards and identical hyperparameters; only the GRPO training seed differs.}
\label{tab:supp-seeds}
\begin{tabular}{lcc}
\toprule
Seed & F1 & Accuracy [95\% CI] \\
\midrule
3407 (default) & 0.795 & 0.660 [0.635, 0.686] \\
1234 & 0.789 & 0.652 [0.626, 0.678] \\
2025 & 0.771 & 0.627 [0.601, 0.654] \\
\midrule
Mean $\pm$ s.d. & $0.785 \pm 0.013$ & $0.646 \pm 0.017$ \\
\bottomrule
\end{tabular}
\end{table}

The three-seed mean F1 is $0.785 \pm 0.013$ with spread 0.024. Two observations follow. First, the diversity ensemble's default-seed lead over the top-5 ranking ensemble (0.795 vs.\ 0.789, gap 0.006) is well inside the seed-to-seed variation, and one of the three seeds (2025, F1 = 0.771) actually trails the default-seed top-5 ensemble. The diversity-vs-ranking comparison is therefore exploratory rather than confirmatory. Second, even the worst seed of the diversity ensemble (F1 = 0.771) exceeds both the three-reward configuration (F1 = 0.738) and the base-rewards-only baseline (F1 = 0.609) by margins much larger than the seed-to-seed spread. The qualitative ordering ``base only $<$ three-reward $<$ five large ensembles'' is robust to seed choice, while finer-grained ranking among the five large ensembles is not.

\section{Strict vs.\ Flexible-Parse Accuracy}
\label{sec:supp-flex}

Both the trained-model accuracy column and the F1 column in the main paper use strict tag-format parsing: any output that does not contain a properly delimited \texttt{<solution>} numeric token is counted as a false negative, even when a numeric answer is present elsewhere in the output. To quantify this format penalty separately from reasoning quality, we recompute accuracy under flexible answer extraction, in which a numeric-token regex over the full completion is used as a fallback when strict extraction fails.

\begin{table}[h]
\centering
\caption{Strict vs.\ flexible-parse accuracy on GSM8K. ``$\Delta$'' is the format penalty borne by the strict metric.}
\label{tab:supp-flex_parse}
\begin{tabular}{lccc}
\toprule
Configuration & Accuracy & Acc.\ (flex) & $\Delta$ \\
\midrule
ens\_diverse\_rounds & 0.660 & 0.673 & +0.012 \\
ens\_top5\_500 & 0.651 & 0.672 & +0.020 \\
ens\_top3 & 0.585 & 0.677 & +0.092 \\
thinking\_steps\_count (best indiv.) & 0.649 & 0.669 & +0.020 \\
\bottomrule
\end{tabular}
\end{table}

For the five large ensembles the format penalty is small (1--2\%). For the three-reward ensemble it is large ($+9.2\%$). Once format compliance is no longer required, \texttt{ens\_top3} is statistically indistinguishable from the larger ensembles. The reading consistent with these numbers is that the three-reward configuration produces correct reasoning but frequently emits a non-canonical solution tag rather than producing wrong reasoning. Under flexible parsing the diversity-vs-ranking gap (0.673 vs.\ 0.672) effectively vanishes.

\section{Reward-Hacking Audit}
\label{sec:supp-hacking}

Three of the top five discovered rewards count structural features of the \texttt{<thinking>} block without verifying answer correctness. \texttt{thinking\_steps\_count} (F1 = 0.787) is the most aggressive: it pays $+3.0$ for at least three newline-separated lines and is the textbook target for length-padding hacks. We therefore audit the policy trained with this reward.

On the 1{,}319 GSM8K test items the trained policy emits a mean of 6.13 lines inside \texttt{<thinking>} (median 6, $p_{10}/p_{90} = 3/10$). The $\geq 3$ threshold is exceeded organically rather than minimally. Splitting the same outputs by correctness, correct responses contain 6.47 lines on average and incorrect responses 5.51, and the fraction of lines that match an arithmetic-step pattern (\verb|\d+\s*[+\-*/]\s*\d+|) is 0.63 for correct vs.\ 0.51 for incorrect. If the model were padding empty lines to satisfy the structural threshold, the correct vs.\ incorrect distributions would be similar or inverted. Instead, correct answers are accompanied by both more lines and a higher density of arithmetic content. The simplest line-padding hypothesis is therefore not supported by the per-question evidence. This audit does not rule out subtler hacks: it shows only that the structural reward is not fully gameable in the most obvious way under this base model and budget.

\section{Computational Cost}
\label{sec:supp-cost}

Individual reward trials take approximately 40 to 50 minutes of wall-clock time per candidate, with variation driven by the length of generated completions. The entire search over 5 rounds and 50 rewards takes approximately 40 hours of wall-clock time on a single NVIDIA RTX 5090 GPU (32 GB VRAM). There is no clear correlation between training time and reward quality; the effectiveness of a reward function depends on its design rather than the computational resources consumed.

\section{Additional Figures}
\label{sec:supp-figures}

\begin{figure}[h]
\centering
\includegraphics[width=0.9\textwidth]{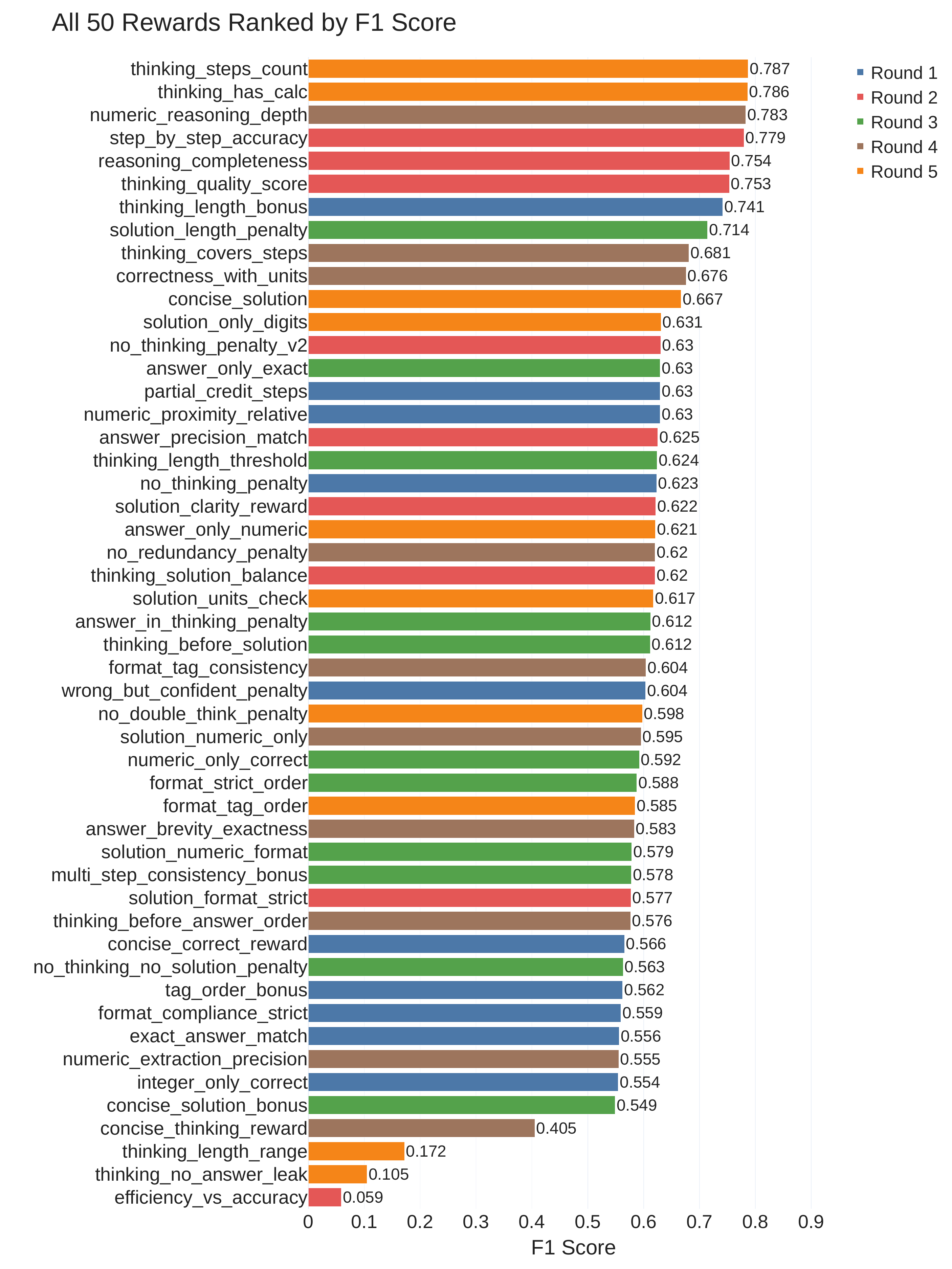}
\caption{F1 scores of all 50 individual reward functions, ranked from highest to lowest, colored by round. Top-performing rewards tend to come from later rounds.}
\label{fig:supp-f1-ranking}
\end{figure}

\bibliographystyle{ieeetr}
\bibliography{references}

\begin{IEEEbiography}[{\includegraphics[width=1in,height=1.25in,clip,keepaspectratio]{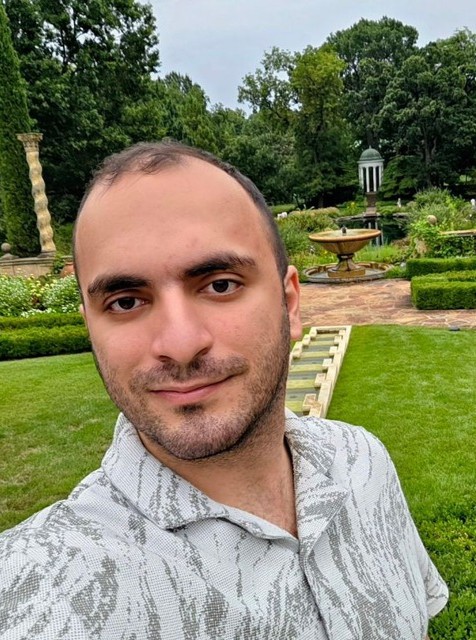}}]{Arash Ahmadi}\,received the B.S. degree in computer engineering from the University of Kurdistan in 2023. He is currently pursuing the M.S. and Ph.D. degrees in electrical and computer engineering at the University of Oklahoma, Norman, OK, USA, where he is a Graduate Research Assistant with the INQUIRE Laboratory under the supervision of Dr. Yaser M. Banad and Dr. Sarah S. Sharif. His research focuses on reinforcement learning, parameter efficient fine tuning of language models, and the deployment of small language models on resource constrained edge devices for agentic systems.
\end{IEEEbiography}

\begin{IEEEbiography}[{\includegraphics[width=1in,height=1.25in,clip,keepaspectratio]{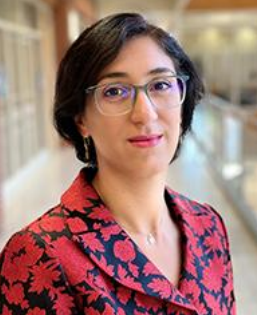}}]{Sarah S. Sharif}\,(Senior Member, IEEE) is an assistant professor in the School of Electrical and Computer Engineering at the University of Oklahoma. She holds a Ph.D. in Electrical Engineering and two M.Sc.\ degrees in Physics and Electrical Engineering, with 10+ years of industrial R\&D experience. Her work in the INQUIRE Laboratory focuses on next generation computing and sensing through optical, quantum optical, and neuromorphic devices.
\end{IEEEbiography}

\begin{IEEEbiography}[{\includegraphics[width=1in,height=1.25in,clip,keepaspectratio]{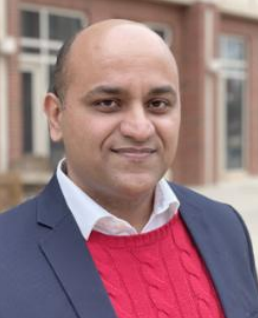}}]{Yaser M. Banad}\,(Senior Member, IEEE) is an Associate Professor in the School of Electrical and Computer Engineering at the University of Oklahoma. He received the M.Sc.\ and Ph.D.\ degrees in electrical and computer engineering from Louisiana State University in 2016. An author of over 100 peer-reviewed publications, his research spans neuromorphic computing, energy-efficient devices, neural-inspired AI acceleration, and semiconductor materials.
\end{IEEEbiography}

\end{document}